\documentclass[10pt, conference, compsocconf]{IEEEtran}
\usepackage{times}
\usepackage{latexsym}
\usepackage{amsmath}
\usepackage{url}
\usepackage{helvet}
\usepackage{courier}
\usepackage{caption}
\usepackage{subcaption}
\usepackage{mathtools}
\usepackage{multirow}
\usepackage{algorithm}
\usepackage{algorithmic}
\usepackage{graphicx}
\graphicspath{ {./images/} }
\begin{document}

\title{Unsupervised Terminological Ontology Learning \\ based on Hierarchical Topic Modeling}

\author{\IEEEauthorblockN{Xiaofeng Zhu\IEEEauthorrefmark{1},
Diego Klabjan\IEEEauthorrefmark{2},
and
Patrick N Bless\IEEEauthorrefmark{3}}
\IEEEauthorblockA{\IEEEauthorrefmark{1}EECS,
Northwestern University, Evanston, IL, USA\\ Email: xiaofengzhu2013@u.northwestern.edu}
\IEEEauthorblockA{\IEEEauthorrefmark{2}IEMS, Northwestern University, Evanston, IL, USA\\
Email: d-klabjan@northwestern.edu}
\IEEEauthorblockA{\IEEEauthorrefmark{3}Intel Corporation, Chandler, AZ, USA\\
Email: patrick.n.bless@intel.com}}

\maketitle

\begin{abstract}
   In this paper, we present \textbf{h}ierarchical \textbf{r}elation-based \textbf{l}atent \textbf{D}irichlet \textbf{a}llocation (hrLDA), a data-driven hierarchical topic model for extracting terminological ontologies from a large number of heterogeneous documents. In contrast to traditional topic models, hrLDA relies on noun phrases instead of unigrams, considers syntax and document structures, and enriches topic hierarchies with topic relations. Through a series of experiments, we demonstrate the superiority of hrLDA over existing topic models, especially for building hierarchies. Furthermore, we illustrate the robustness of hrLDA in the settings of noisy data sets, which are likely to occur in many practical scenarios. Our ontology evaluation results show that ontologies extracted from hrLDA are very competitive with the ontologies created by domain experts. 
\end{abstract}

%-------------------------------------------------------------------------
\section{Introduction}
\label{Introduction} 
Although researchers have made significant progress on knowledge acquisition and have proposed many ontologies, for instance, WordNet \cite{Miller1995}, DBpedia \cite{Bizer07dbpedia:a}, YAGO \cite{Suchanek07yago:a}, Freebase, \cite{Bollacker08freebase:a} Nell \cite{Carlson10towardan}, DeepDive \cite{niu2012deepdive}, Domain Cartridge \cite{mukherjee2014domain}, Knowledge Vault \cite{dong2014knowledge}, INS-ES \cite{Wei:2015:LKB:2806416.2806513}, iDLER \cite{chatzis2015inducing}, and TransE-NMM \cite{nguyen2016neighborhood}, current ontology construction methods still rely heavily on manual parsing and existing knowledge bases. This raises challenges for learning ontologies in new domains. While a strong ontology parser is effective in small-scale corpora, an unsupervised model is beneficial for learning new entities and their relations from new data sources, and is likely to perform better on larger corpora.

\begin{figure}[h!]  
\centering
  \includegraphics[width= 0.48\textwidth]{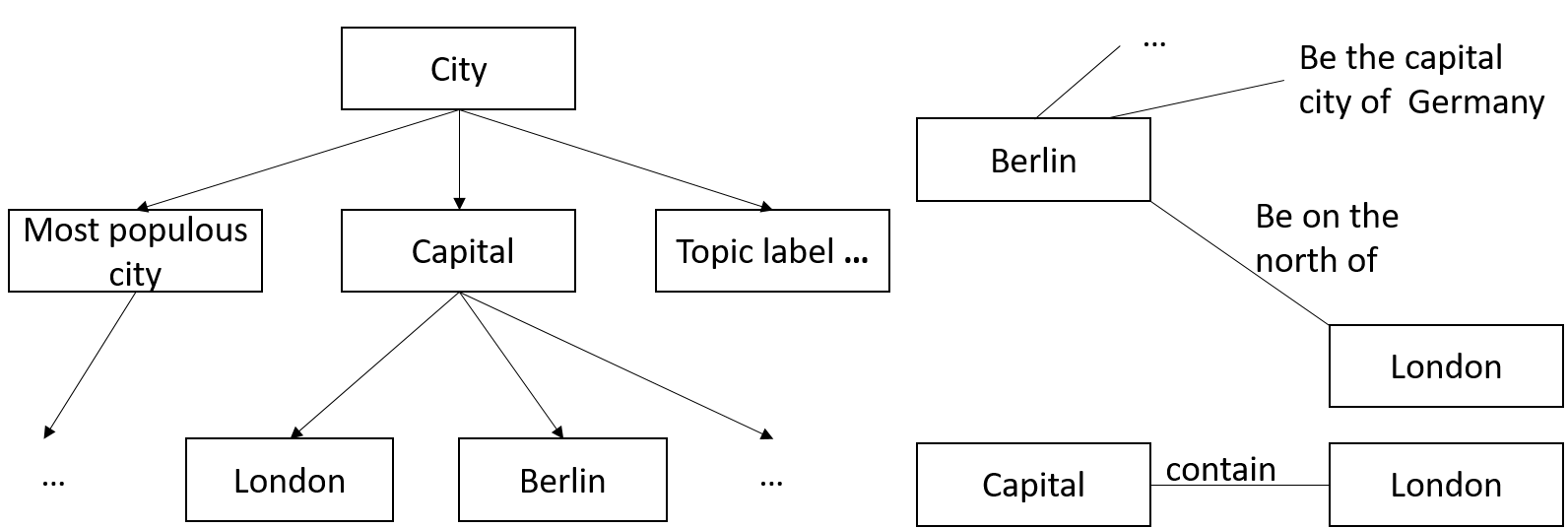}
  \caption{A representation of a terminological ontology. (Left: topic hierarchies) Topic $city$ is composed of $most$ $populous$ $city$, $capital$, $London$, $Berlin$, etc. $City$ $\rightarrow$ $capital$ $\rightarrow$ $London$ and $city$ $\rightarrow$ $capital$ $\rightarrow$ $Berlin$ are two topic paths. (Right: topic relations) Every topic label has relations to itself and/or with other labels. $Be$ $the$ $capital$ $city$ $of$ $Germany$ is one relation/property of topic $Berlin$. $Be$ $on$ $the$ $north$ $of$ is one relation of topic $Berlin$ to $London$.}
  \label{fig:terminological_ontology}
   
\end{figure}

In this paper, we focus on unsupervised terminological ontology learning and formalize a terminological ontology as a hierarchical structure of subject-verb-object triplets. We divide a terminological ontology into two components: \textbf{topic hierarchies} and \textbf{topic relations}. Topics are presented in a tree structure where each node is a topic label (noun phrase), the root node represents the most general topic, the leaf nodes represent the most specific topics, and every topic is composed of its topic label and its descendant topic labels. Topic hierarchies are preserved in topic paths, and a topic path connects a list of topics labels from the root to a leaf. Topic relations are semantic relationships between any two topics or properties used to describe one topic. Figure \ref{fig:terminological_ontology} depicts an example of a terminological ontology learned from a corpus about European cities. We extract terminological ontologies by applying unsupervised hierarchical topic modeling and relation extraction to plain text.

Topic modeling was originally used for topic extraction and document clustering. The classical topic model, latent Dirichlet allocation (LDA)~\cite{Blei03latentdirichlet}, simplifies a document as a bag of its words and describes a topic as a distribution of words. Prior research \cite{ocampo2009data, Wang2010, jing2012ontology, slutsky2013tree, colace2014terminological, movshovitz2015kb, hu2016grounding} has shown that LDA-based approaches are adequate for (terminological) ontology learning. However, these models are deficient in that they still need human supervision to decide the number of topics, and to pick meaningful topic labels usually from a list of unigrams.
Among models not using unigrams, LDA-based Global Similarity Hierarchy Learning (LDA+GSHL) \cite{Wang2010} only extracts a subset of relations: ``broader" and ``related" relations. In addition, the topic hierarchies of KB-LDA \cite{movshovitz2015kb} rely on hypernym-hyponym pairs capturing only a subset of hierarchies.

Considering the shortcomings of the existing methods, the main objectives of applying topic modeling to ontology learning are threefold.
   \begin{enumerate}
     \item In topic models, a topic is usually represented with a list of unigrams. In a terminological ontology, a topic/entity needs to be represented with a more descriptive identifier (i.e., noun phrase). Currently, the number of topics is usually a fixed parameter, which restricts the number of classes an ontology could have. For instance, it is difficult to add a new species to an animal ontology.
     \item Both relations among different noun phrases and relations/properties (see the relations in Figure \ref{fig:terminological_ontology}) for describing single noun phrases should be captured during the topic generation process.
     \item Hierarchies need to be built on topical affiliations. If topic $B$ is a sub-topic of topic $A$, $B$ has a more specific meaning than $A$. The depth of each topic path should be determined by a data-driven method.
   \end{enumerate}

To achieve the first objective, we extract noun phrases and then propose a sampling method to estimate the number of topics. For the second objective, we use language parsing and relation extraction to learn relations for the noun phrases. Regarding the third objective, we adapt and improve the hierarchical latent Dirichlet allocation (hLDA) model \cite{Blei04hierarchicaltopic,blei2010nested}. hLDA is not ideal for ontology learning because it builds topics from unigrams (which are not descriptive enough to serve as entities in ontologies) and the topics may contain words from multiple domains when input data have documents from many domains (see Section \ref{Background} and Figure \ref{fig: noisy results}). Our model, hrLDA, overcomes these deficiencies. In particular, hrLDA represents topics with noun phrases, uses syntax and document structures such as paragraph indentations and item lists, assigns multiple topic paths for every document, and allows topic trees to grow vertically and horizontally. 
%Similar to LDA+GSHL, hrLDA creates a topic tree where every node is a noun phrase. This is different from the topic trees generated by hLDA and KB-LDA where every node is a topic, which is a list of terms.

The primary contributions of this work can be specified as follows.
   \begin{itemize}
     \item We develop a hierarchical topic model, hrLDA, that does not require one to set the topic number at every level of a topic tree or to set the topic path lengths from the root to leaves.
     \item We integrate relation extraction into topic modeling leading to lower perplexity.
     \item We propose a multiple topic path drawing strategy, which is an improvement over the simple topic path drawing method proposed in hLDA. 	
     \item We present automatic extraction of terminological ontologies via hrLDA.
     \end{itemize}

The rest of this paper is organized into five parts. In Section 2, we provide a brief background of hLDA. In Section 3, we present our hrLDA model and the ontology generation method. In Section 4, we demonstrate empirical results regarding topic hierarchies and generated terminological ontologies. Finally, in Section 5, we present some concluding remarks and discuss avenues for future work and improvements.

\section{Background}
\label{Background}
In this section, we introduce our main baseline model, hierarchical latent Dirichlet allocation (hLDA), and some of its extensions. We start from the components of hLDA - latent Dirichlet allocation (LDA) and the Chinese Restaurant Process (CRP)- and then explain why hLDA needs improvements in both building hierarchies and drawing topic paths.

LDA is a three-level Bayesian model in which each document is a composite of multiple topics, and every topic is a distribution over words. 
Due to the lack of determinative information, LDA is unable to distinguish different instances containing the same content words, (e.g. ``I trimmed my polished \underline{nails}" and ``I have just hammered many rusty \underline{nails}"). In addition, in LDA all words are probabilistically independent and equally important. This is problematic because different words and sentence elements should have different contributions to topic generation. For instance, articles contribute little compared to nouns, and sentence subjects normally contain the main topics of a document.

Introduced in hLDA, CRP partitions words into several topics by mimicking a process in which customers sit down in a Chinese restaurant with an infinite number of tables and an infinite number of seats per table. Customers enter one by one, with a new customer choosing to sit at an occupied table or a new table. The probability of a new customer sitting at the table with the largest number of customers is the highest. In reality, customers do not always join the largest table but prefer to dine with their acquaintances. The theory of distance-dependent CRP was formerly proposed by David Blei \cite{blei2011distance}. We provide later in Section \ref{Acquaintance Chinese Restaurant Process} an explicit formula for topic partition given that adjacent words and sentences tend to deal with the same topics.

hLDA combines LDA with CRP by setting one topic path with fixed depth $L$ for each document. The hierarchical relationships among nodes in the same path depend on an $L$ dimensional Dirichlet distribution that actually arranges the probabilities of topics being on different topic levels. Despite the fact that the single path was changed to multiple paths in some extensions of hLDA - the nested Chinese restaurant franchise processes \cite{ahmed2013nested} and the nested hierarchical Dirichlet Processes \cite{paisley2015nested}, - this topic path drawing strategy puts words from different domains into one topic when input data are mixed with topics from multiple domains. This means that if a corpus contains documents in four different domains, hLDA is likely to include words from the four domains in every topic (see Figure \ref{fig: noisy results}). In light of the various inadequacies discussed above, we propose a relation-based model, hrLDA. hrLDA incorporates semantic topic modeling with relation extraction to integrate syntax and has the capacity to provide comprehensive hierarchies even in corpora containing mixed topics.

\section{Hierarchical Relation-based Latent Dirichlet Allocation}
The main problem we address in this section is generating terminological ontologies in an unsupervised fashion. The fundamental concept of hrLDA is as follows. When people construct a document, they start with selecting several topics. Then, they choose some noun phrases as subjects for each topic. Next, for each subject they come up with relation triplets to describe this subject or its relationships with other subjects. Finally, they connect the subject phrases and relation triplets to sentences via  reasonable grammar. The main topic is normally described with the most important relation triplets. Sentences in one paragraph, especially adjacent sentences, are likely to express the same topic. 

We begin by describing the process of reconstructing LDA. Subsequently, we explain relation extraction from heterogeneous documents. Next, we propose an improved topic partition method over CRP. Finally, we demonstrate how to build topic hierarchies that bind with extracted relation triplets.  

\subsection{Relation-based Latent Dirichlet Allocation}
Documents are typically composed of chunks of texts, which may be referred to as sections in Word documents, paragraphs in PDF documents, slides in presentation documents, etc. Each chunk is composed of multiple sentences that are either atomic or complex in structure, which means a document is also a collection of atomic and/or complex sentences. An atomic sentence (see module $T$ in Figure  \ref{fig:rLDA}) is a sentence that contains only one subject ($S$), one object ($O$) and one verb ($V$) between the subject and the object. For every atomic sentence whose object is also a noun phrase, there are at least two relation triplets (e.g., ``\textit{The tiger that gave the excellent speech is handsome}" has relation triplets: (\textit{tiger, give, speech}), {(\textit{speech, be given by, tiger}), and (\textit{tiger, be, handsome})). By contrast, a complex sentence can be subdivided into multiple atomic sentences. Given that the syntactic verb in a relation triplet is determined by the subject and the object, a document $d$ in a corpus $D$ can be ultimately reduced to $N_d$ subject phrases (we convert objects to subjects using passive voice) associated with $N_d$ relation triplets $T_d$. Number $N_d$ is usually larger than the actual number of noun phrases in document $d$. By replacing the unigrams in LDA with relation triplets, we retain definitive information and assign salient noun phrases high weights. 

We define $Dir(\alpha)$ as a Dirichlet distribution parameterized by hyperparameters $\alpha$, $Multi(\theta)$ as a multinomial distribution parameterized by hyperparameters $\theta$, $Dir(\eta)$ as a Dirichlet distribution parameterized by $\eta$, and $Multi(\beta)$ as a multinomial distribution parameterized by $\beta$. We assume the corpus has $K$ topics. Assigning $K$ topics to the $N_d$ relation triplets of document $d$ follows a multinomial distribution $Multi(\theta)$ with prior $Dir(\alpha)$. Selecting the $N_d$ relation triplets for document $d$ given the $K$ topics follows a multinomial distribution $Multi(\beta)$ with prior $Dir(\eta)$. We denote $T=\{T_d\}_{d\in D}$ as the list of relation triplet lists extracted from all documents in the corpus, and $Z$ as the list of topic assignments of $T$. We denote the relation triplet counts of documents in the corpus by $N=\{N_d\}_{d\in D}$. The graphical representation of the relation-based latent Dirichlet allocation (rLDA) model is illustrated in Figure \ref{fig:rLDA}.

\begin{figure}[h!]  
\centering
  \includegraphics[width= 0.48\textwidth]{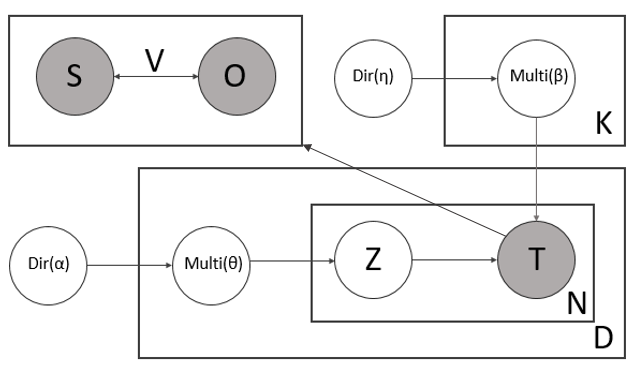}
  \caption{Plate notation of rLDA}
  \label{fig:rLDA}
     
\end{figure}

The plate notation can be decomposed into two types of Dirichlet-multinomial conjugated structures: document-topic distribution $Dir(\alpha) \rightarrow Multi(\theta) \rightarrow Z $ and topic-relation distribution $Dir(\eta) \rightarrow Multi(\beta) \rightarrow T|Z$. Hence, the joint distribution of $T$ and $Z$ can be represented as
\begin{equation}
\label{equation:rLDA}
\begin{split}
	&P(T, Z | \alpha, \eta) \\
	&=P(T | Z, \eta) \  P(Z | \alpha) \\
	&= {\prod_{k = 1}^K \displaystyle \frac{Dir(C_k + \eta)}{Dir(\eta)}} {\prod_{d}^D \displaystyle \frac{Dir(B_d + \alpha)}{Dir(\alpha)}} \\	
	%Dir(\bigstar) &=  \int \prod_{i = 1} p_i^{\bigstar_i - 1} dp \\
&C_k = (C_k^1,C_k^2, ..., C_k^w, ..., C_k^W ) \\
&B_d = (B_d^1,B_d^2, ..., B_d^k, ..., B_d^K ), \\	
\end{split}
\end{equation}
where $W$ is the number of unique relations in all documents, $C_k^w$ is the number of occurrences of the relation triplet $w$ generated by topic $k$ in all documents, and $B_d^k$ is the number of relation triplets generated by topic $k$ in document $d$. $Dir(\alpha)$ is a conjugate prior for $Multi(\theta)$ and thus the posterior distribution is a new Dirichlet distribution parameterized by $(B_d + \alpha)$. The same rule applies to $Dir(C_k + \eta)$.

\subsection{Relation Triplet Extraction}
Extracting relation triplets is the essential step of hrLDA, and it is also the key process for converting a hierarchical topic tree to an ontology structure. The idea is to find all syntactically related noun phrases and their connections using a language parser such as the Stanford NLP parser \cite{manning-EtAl:2014:P14-5} and Ollie \cite{ollie-emnlp12}. Generally, there are two types of relation triplets:
% relation examples
   \begin{itemize}
     \item Subject-predicate-object-based
      relations,\\
     e.g., \textit{New York is the largest city in the United States} $\Rightarrow$ (\textit{New York, be the largest city in, the United States});
     \item Noun-based/hidden relations,\\
      e.g., \textit{Queen Elizabeth} $\Rightarrow$ (\textit{Elizabeth, be, queen}). \\
%      and, \textit{Joint Electron Device Engineering Council (JEDEC)}\\ $\Rightarrow$ (\textit{JEDEC, be short for, Joint Electron Device Engineering Council}).      
   \end{itemize}
A special type of relation triplets can be extracted from presentation documents such as those written in PowerPoint using document structures. Normally lines in a slide are not complete sentences, which means language parsing does not work. However, indentations and bullet types usually express inclusion relationships between adjacent lines. Starting with the first line in an itemized section, our algorithm scans the content in a slide line by line, and creates relations based on the current item and the item that is one level higher. 

\subsection{Acquaintance Chinese Restaurant Process}
\label{Acquaintance Chinese Restaurant Process}
As mentioned in Section 2, CRP always assigns the highest probability to the largest table, which assumes customers are more likely to sit at the table that has the largest number of customers. This ignores the social reality that a person is more willing to choose the table where his/her closest friend is sitting even though the table also seats unknown people who are actually friends of friends. Similarly with human-written documents, adjacent sentences usually describe the same topics. We consider a restaurant table as a topic, and a person sitting at any of the tables as a noun phrase. In order to penalize the largest topic and assign high probabilities to adjacent noun phrases being in the same topics, we introduce an improved partition method, Acquaintance Chinese Restaurant Process (ACRP). 

The ultimate purposes of ACRP are to estimate $K$, the number of topics for rLDA, and to set the initial topic distribution states for rLDA. Suppose a document is read from top to bottom and left to right. As each noun phrase belongs to one sentence and one text chunk (e.g., section, paragraph and slide), the locations of all noun phrases in a document can be mapped to a two-dimensional space where \textbf{sentence location} is the x axis and text \textbf{chunk location} is the y axis (the first noun phrase of a document holds value (0, 0)). More specifically, every noun phrase has four attributes: \textbf{content}, \textbf{location}, \textbf{one-to-many relation triplets}, and \textbf{document ID}. Noun phrases in the same text chunk are more likely to be ``acquaintances;" they are even closer to each other if they are in the same sentence. In contrast to CRP, ACRP assigns probabilities based on closeness, which is specified in the following procedure.

% ACRP algorithm
\begin{enumerate}
   \item Let $z_{n}$ be the integer-valued random variable corresponding to the index of a topic assigned to the $n^{th}$ phrase. Draw a probability $P(z_{n+1})$ from Equations \ref{eq: first unoccupied} to \ref{eq: otherwise} below for the $(n+1)^{th}$ noun phrase $t^{n+1}$, joining each of the existing $k$ topics and the new $(k+1)^{th}$ topic given the topic assignments of previous $n$ noun phrases, $Z_{1:n}$. If a noun phrase joins any of the existing k topics, we denote the corresponding topic index by $i \in [1, k]$.
   \begin{itemize}
     \item The probability of choosing the $(k + 1)^{th}$ topic:
      \begin{equation}
      \label{eq: first unoccupied}
      P(z_{n+1} = (k+1) | Z_{1:n}) = \frac{\gamma}{n+ \gamma}.
      \end{equation}
     \item The probability of selecting any of the $k$ topics:
     	\begin{itemize}
      	\item if the content of $t^{n+1}$ is synonymous with or an acronym of a previously analyzed noun phrase $t^m$ $(m < n +1)$ in the $i^{th}$ topic,
      	\begin{equation}
      	P(z_{n+1} = i | Z_{1:n}) = 1 - \gamma;
      	\end{equation}
       \item else if the document ID of $t^{n+1}$ is different from all document IDs belonging to the $i^{th}$ topic,     
      	\begin{equation}
     	 P(z_{n+1} = i | Z_{1:n}) = \gamma;
      	\end{equation}
	 	 \item otherwise,
     	\begin{equation}
     	\label{eq: otherwise}
     	\begin{split}
     		&P(z_{n+1} = i | Z_{1:n}) = \\
     		&\frac{C_i - (1 - \frac{1}{min(Q_{1:i})})}{(1 + min(S_{1:i})) n+ \gamma},
     	\end{split}
     	\end{equation}
     	where $C_i$ refers to the current number of noun phrases in the $i^{th}$ topic, $Q_{1:i}$ represents the vector of \textbf{chunk location} differences of the $(n +1)^{th}$ noun phrase and all members in the $i^{th}$ topic, $S_{1:i}$ stands for the vector of \textbf{sentence location} differences, and $\gamma$ is a penalty factor.
    	\end{itemize}         
	\end{itemize}
Normalize the ($k + 1$) probabilities to guarantee they are each in the range of [0, 1] and their sum is equal to 1.
   \item Based on the probabilities \ref{eq: first unoccupied} to \ref{eq: otherwise}, we sample a topic index $z$ from $\{1, ..., (k + 1)\}$ for every noun phrase, and we count the number of unique topics $K$ in the end. We shuffle the order of documents and iterate ACRP until $K$ is unchanged.
\end{enumerate} 

\subsection{Nested Acquaintance Chinese Restaurant Process}
\label{Nested Acquaintance Chinese Restaurant Process}
The procedure for extending ACRP to hierarchies is essential to why hrLDA outperforms hLDA. Instead of a predefined tree depth $L$, the tree depth for hrLDA is optional and data-driven. More importantly, clustering decisions are made given a global distribution of all current non-partitioned phrases (leaves) in our algorithm. This means there can be multiple paths traversed down a topic tree for each document. With reference to the topic tree, every node has a noun phrase as its label and represents a topic that may have multiple sub-topics. The root node is visited by all phrases. In practice, we do not link any phrases to the root node, as it contains the entire vocabulary. An inner node of a topic tree contains a selected topic label. A leaf node contains an unprocessed noun phrase. We define a hashmap $leaves$ with a document ID as the key and the current leaf nodes of the document as the value. We denote the current tree level by $l$. We next outline the overall algorithm.

% hrLDA algorithm
\begin{enumerate}
   \item We start with the root node ($l = 0$) and apply rLDA to all the documents in a corpus.
   \begin{enumerate}
     \item Collect the current leaf nodes of every document. $leaves$ initially contains all noun phrases in the corpus. Assign a cluster partition to the leaf nodes in each document based on ACRP and sample the cluster partition until the number of topics of all noun phrases in $leaves$ is stable or the iteration reaches the predefined number of iteration times (whichever occurs first).
     \item Mark the number of topics (child nodes) of parent node $m$ at level $l$ as $K^{l_m}$. Build a $K^{l_m}$- dimensional topic proportion vector $\theta$ based on $Dir(\alpha)$.
     \item For every noun phrase $\{t_n\}_{n=1}^{N_d}$ in document $d$, form the topic assignments $Z_{\lbrace 1, . . . , K_{l_m} \rbrace}$ based on $Multi(\theta)$.
     \item Generate relation triplets from $Multi(\beta)$ given $Dir(\eta)$ and the associated topic vector $\{Z_k\}_{k=1}^{K^{l_m}}$. 
     \item Eliminate partitioned leaf nodes from $leaves$. Update the current level $l$  by $1$.
   \end{enumerate}
   \item If phrases in $leaves$ are not yet completely partitioned to the next level and $l$ is less than $L$, continue the following steps. For each leaf node, we set the top phrase (i.e., the phrase having the highest probability) as the topic label of this leaf node and the leaf node becomes an inner node. We next update $leaves$ and repeat procedures $1 (a) - 1 (e)$.
\end{enumerate}

To summarize this process more succinctly: we build the topic hierarchies with rLDA in a divisive way (see Figure \ref{fig:hrLDA}). We start with the collection of extracted noun phrases and split them using rLDA and ACRP. Then, we apply the procedure recursively until each noun phrase is selected as a topic label. After every rLDA assignment, each inner node only contains the topic label (top phrase), and the rest of the phrases are divided into nodes at the next level using ACRP and rLDA. Hence, we build a topic tree with each node as a topic label (noun phrase), and each topic is composed of its topic labels and the topic labels of the topic's descendants. In the end, we finalize our terminological ontology by linking the extracted relation triplets with the topic labels as subjects.

\begin{figure}[h!]  
\centering
  \includegraphics[width= 0.53\textwidth]{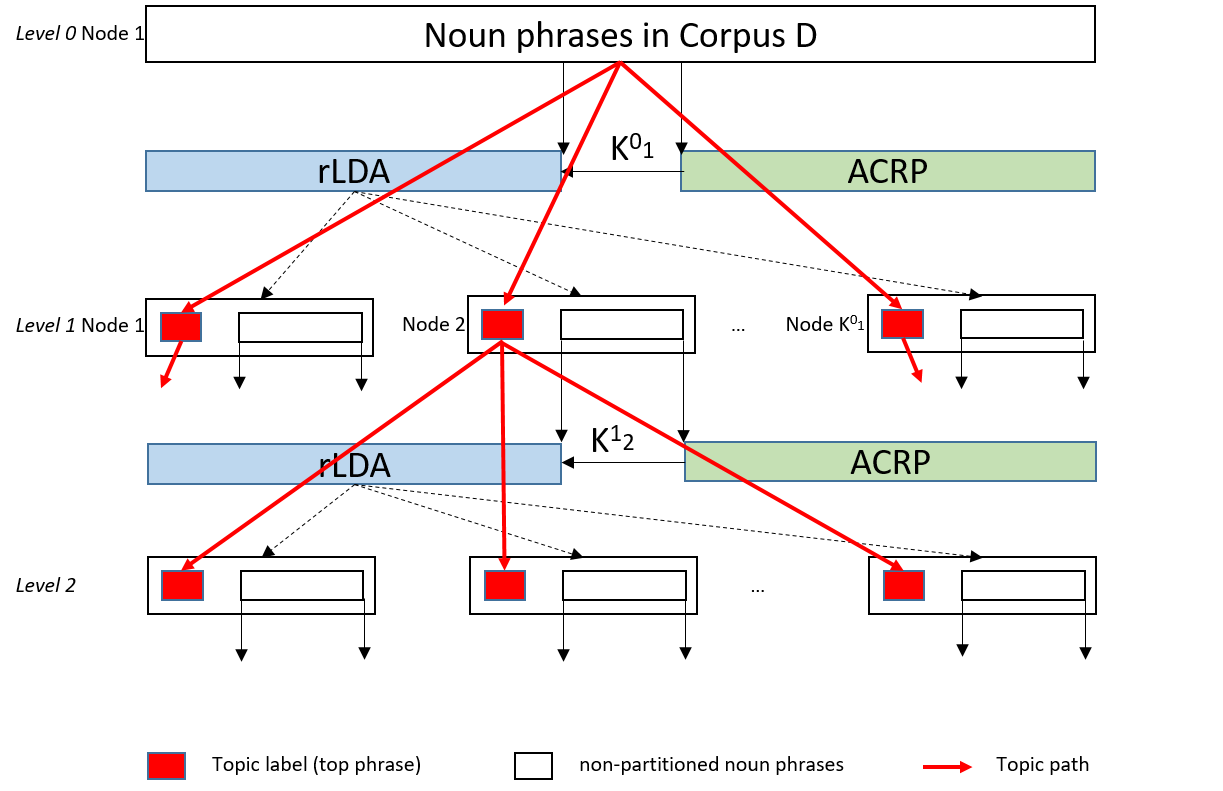}
  \caption{Graphical representation of hrLDA}
  \label{fig:hrLDA}
   
\end{figure}
We use collapsed Gibbs sampling \cite{griffiths2004finding} for inference from posterior distribution $P(Z|T, \alpha, \eta)$ based on Equation \ref{equation:rLDA}. Assume the $n^{th}$ noun phrase $t_n = \hat{t}$ in parent node $m$ comes from document $d$. We denote unassigned noun phrases from document $d$ in parent node $m$ by $d^m$, and unique noun phrases in parent node $m$ by $\hat{T}_m$. We simplify the probability of assigning the $n^{th}$ noun phrase in parent node $m$ to topic $k$ among $K^{l_m}$ topics as
\begin{equation}
\begin{split}
	& P(z_{n} = k | Z_{\neg n}, \hat{T}_m, \alpha, \eta) \\ 
	& \propto P(t_n = \hat{t}, z_{n} = k | Z_{\neg n}, \hat{T}_{m{\neg n}}, \alpha, \eta) \\
	& = \int P(t_n = \hat{t}, z_{n} = k | Z_{\neg n}, \hat{T}_{m{\neg n}}, \theta_{d^m}, \beta_k) d\theta_{d^m}, d\beta_k\\
	& = \frac{C_{k, \hat{t}\neg n} + \eta}{\sum_{\hat{t}}^{\hat{T}_m} (C_{k, \hat{t} \neg n} + \eta)} \frac{C_{d^m, k\neg n} + \alpha}{\sum_{k=1}^{K^{l_m}}(C_{d^m, k\neg n} + \alpha)} \\ 
	%& = \frac{C_{k, \hat{t} \neg n} + \eta}{\sum_{\hat{t}}^{\hat{T}_m} (C_{k, \hat{t} \neg n} + \eta)} C_{d^m, k\neg n} + \alpha,
\end{split}
\end{equation}
where $Z_{\neg n}$ refers to all topic assignments other than $z_{n}$, $\theta_{d^m}$ is multinational document-topic distribution for unassigned noun phrases ${d^m}$, $\beta_k$ is the multinational topic-relation distribution for topic $k$, $C_{k, \hat{t}\neg n}$ is the number of occurrences of noun phrase $\hat{t}$ in topic $k$ except the $n^{th}$ noun phrase in $m$, $C_{d^m, k\neg n}$ stands for the number of times that topic $k$ occurs in $d^m$ excluding the $n^{th}$ noun phrase in $m$. The time complexity of hrLDA is $O(\sum_{l=1}^L N^2 K_l)$, where $K_l$ is the number of topics at level $l$. The space complexity is $O(N)$.
%, and $\sum_{t}^{\hat{T}} C_{k, t \neg n}$ holds the number of times that topic $k$ is observed in all noun phrases of parent node $m$ except the $n^{th}$ noun phrase in document $d^m$.

In order to build a hierarchical topic tree of a specific domain, we must generate a subset of the relation triplets using external constraints or semantic seeds via a pruning process \cite{thelen2002bootstrapping}. As mentioned above, in a relation triplet, each relation connects one subject and one object.  By assembling all subject and object pairs, we can build an undirected graph with the objects and the subjects constituting the nodes of the graph \cite{Krause2012}. Given one or multiple semantic seeds as input, we first collect a set of nodes that are connected to the seed(s), and then take the relations from the set of nodes as input to retrieve associated subject and object pairs. This process constitutes one recursive step. The subject and object pairs become the input of the subsequent recursive step.

\section{Empirical Results}
\subsection{Implementation}
We utilized the Apache poi library to parse texts from pdfs, word documents and presentation files; the MALLET toolbox \cite{McCallumMALLET} for the implementations of LDA, optimized\_LDA \cite{Asuncion2009} and hLDA; the Apache Jena library to add relations, properties and members to hierarchical topic trees; and Stanford Protege\footnote{http://protege.stanford.edu/} for illustrating extracted ontologies. We make our code and data available \footnote{https://github.com/XiaofengZhu/hrLDA}. We used the same empirical hyper-parameter setting (i.e., $\alpha = 1$, $\eta = 0.1$, and $\gamma = 0.01$) across all our experiments. We then demonstrate the evaluation results from two aspects: topic hierarchy and ontology rule.

\subsection{Hierarchy Evaluation}
In this section, we present the evaluation results of hrLDA tested against optimized\_LDA, hLDA, and phrase\_hLDA (i.e., hLDA based on noun phrases) as well as ontology examples that hrLDA extracted from real-world text data. The entire corpus we generated contains 349,362 tokens (after removing stop words and cleaning) and is built from articles on $semiconductor$ $packaging$. It includes 84 presentation files, articles from 1,782 Wikipedia pages and 3,000 research papers that were published in IEEE manufacturing conference proceedings within the last decade. In order to see the performance in data sets of different scales, we also used a smaller corpus Wiki that holds the articles collected from the Wikipedia pages only. 

We extract a single level topic tree using each of the four models; hrLDA becomes rLDA, and phrase\_hLDA becomes phrase-based LDA. We have tested the average perplexity and running time performance of ten independent runs on each of the four models \cite{gangopadhyay2012methodology, downey2013using}. Equation \ref{equ:perplexity} defines the perplexity, which we employed as an empirical measure. 
% Perplexity  formula
\begin{equation}
\label{equ:perplexity}
\textstyle ln (perplexity) =  \textstyle - \frac{\sum_{d}^{D} \log(P(T_d | Z_d)P(Z_d|d))}{\sum_{d}^{D} N_d},
%\textstyle ln (perplexity) =  \textstyle - \frac{\sum_{d}^{D} \log(\sum_{k=1}^{K}P(T_d | Z_k)P(Z_k|d))}{\sum_{d}^{D} N_d},
\end{equation}
where $T_d$ is a vector containing the $N_d$ relation triplets in document $d$, and $Z_d$ is the topic assignment for  $T_d$.

The comparison results on our Wiki corpus are shown in Figure \ref{fig:level_1}. hrLDA  yields the lowest perplexity and reasonable running time. As the running time spent on parameter optimization is extremely long (the optimized\_LDA requires 19.90 hours to complete one run), for efficiency, we adhere to the fixed parameter settings for hrLDA. 

% Figure 2
\begin{figure}[h]
    \centering
    \includegraphics[width=0.5\textwidth]{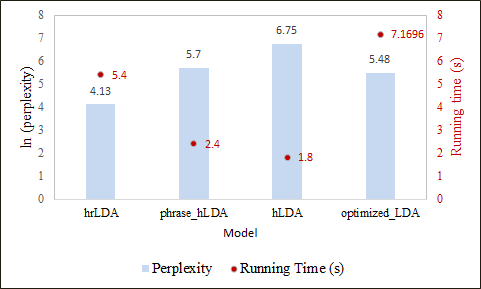}
    \caption{Comparison results of hrLDA, phrase\_hLDA, hLDA and optimized\_LDA on perplexity and running time}
    \label{fig:level_1}
     
\end{figure}

\noindent \textbf{Superiority}

Figures \ref{fig: level 2} to \ref{fig: level 10} illustrates the perplexity trends of the three hierarchical topic models (i.e., hrLDA, phrase\_hLDA and hLDA) applied to both the Wiki corpus and the entire corpus with $seed$ ``\textit{chip}" given different level settings. From left to right, hrLDA retains the lowest perplexities compared with other models as the corpus size grows. Furthermore, from top to bottom, hrLDA  remains stable as the topic level increases, whereas the perplexity of phrase\_hLDA and especially the perplexity of hLDA become rapidly high. Figure \ref{fig: level 10 final} highlights the perplexity values of the three models with confidence intervals in the final state. As shown in the two types of experiments, hrLDA has the lowest average perplexities and smallest confidence intervals, followed by phrase\_hLDA, and then hLDA. 

% Figure 3-7
% Figure 3 level = 2
\begin{figure*}[pt]
\begin{subfigure}[b]{0.5\textwidth}

  \includegraphics[width = \textwidth]{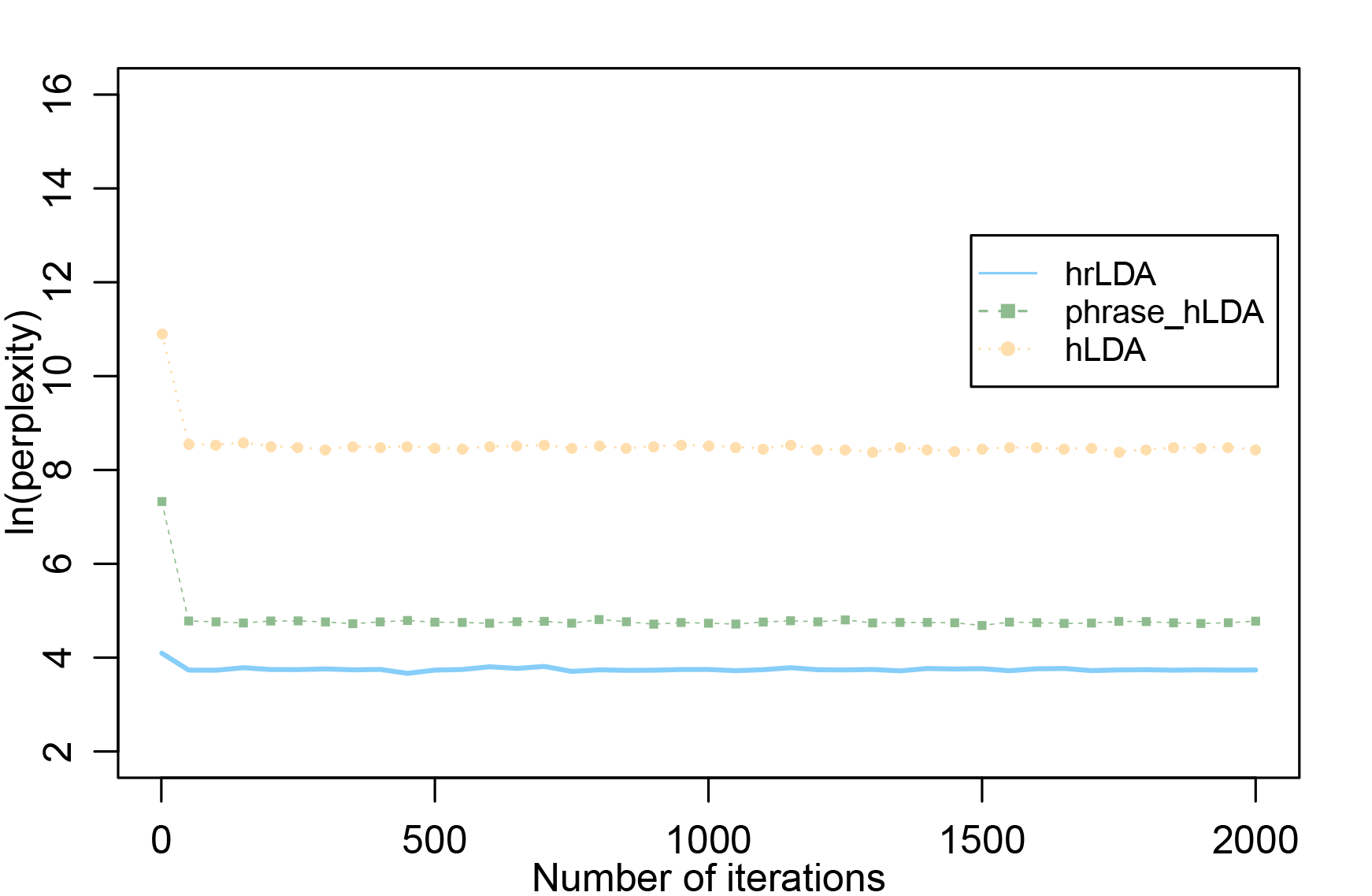}
  \caption{The Wiki corpus}
  \label{fig:Wiki corpus 2}
\end{subfigure}%
\begin{subfigure}[b]{0.5\textwidth}

  \includegraphics[width =\textwidth]{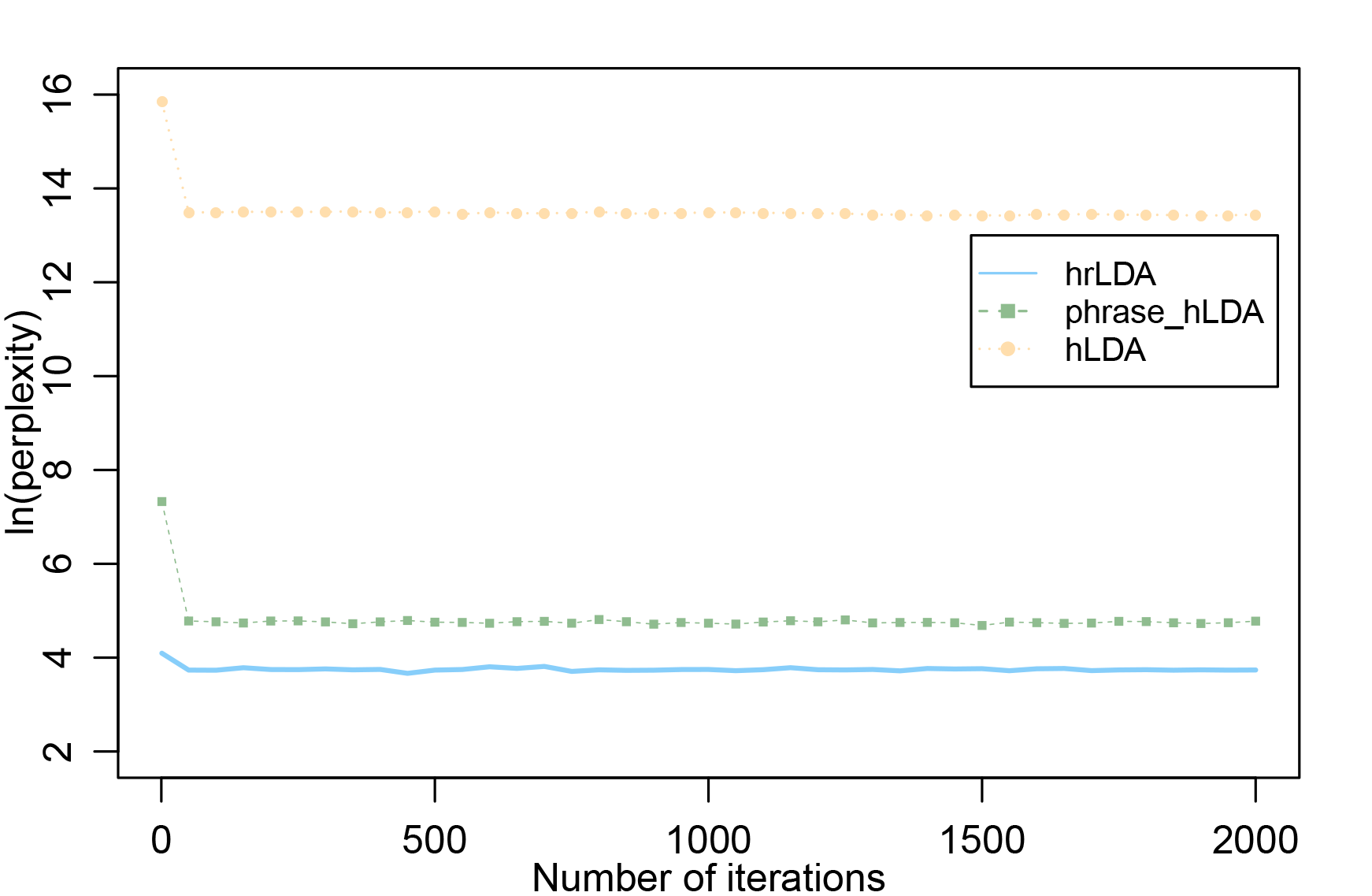}
  \caption{The entire corpus}
  \label{fig:entire corpus 2}%
\end{subfigure}
\caption{Perplexity trends within 2000 iterations with level = 2}
\label{fig: level 2}
\end{figure*}

% Figure 5 level = 6
\begin{figure*}[pt]
\begin{subfigure}[b]{0.5\textwidth}

  \includegraphics[width = \textwidth]{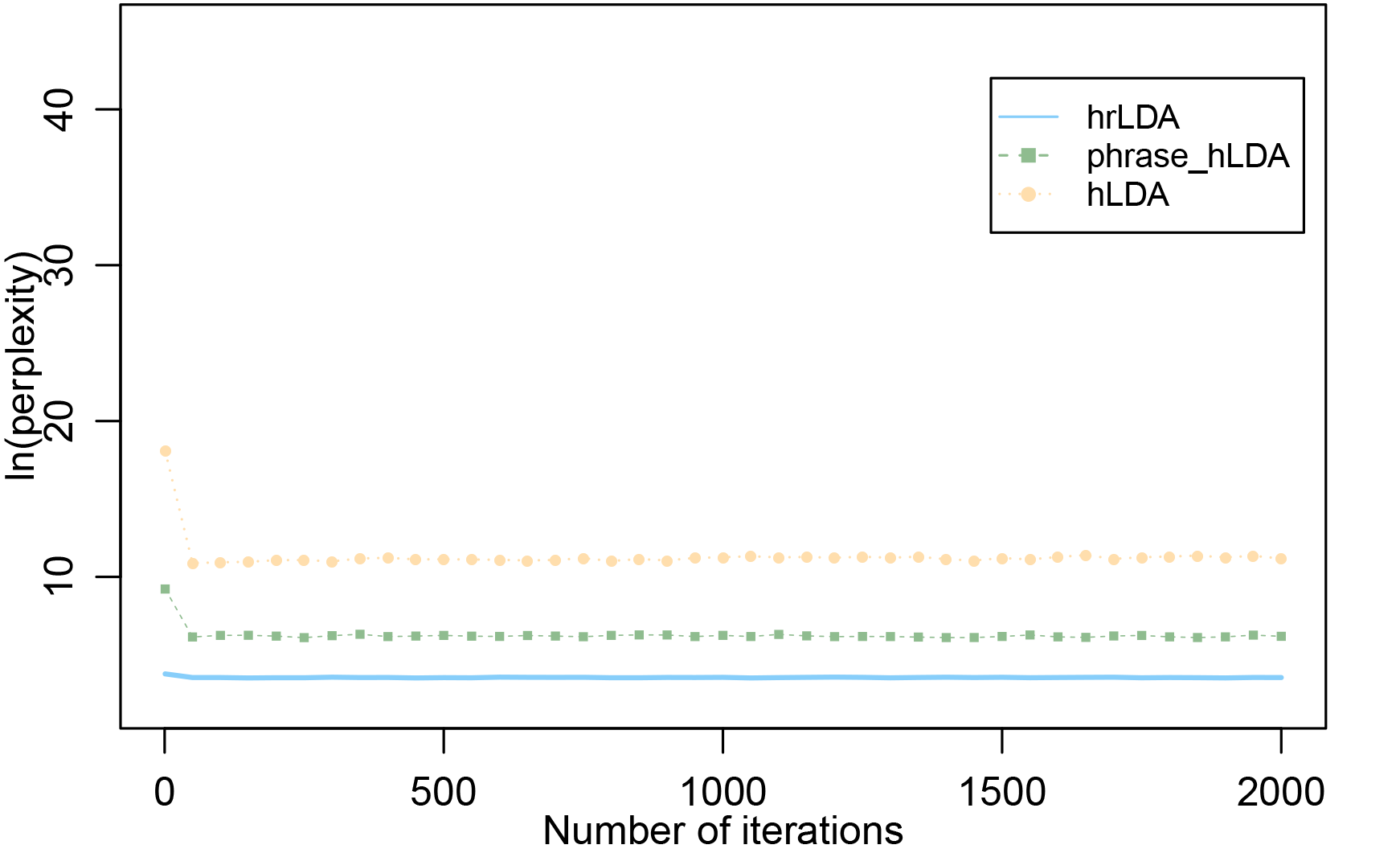}
  \caption{The Wiki corpus}
  \label{fig:Wiki corpus 6}
\end{subfigure}%
\begin{subfigure}[b]{0.5\textwidth}

  \includegraphics[width =\textwidth]{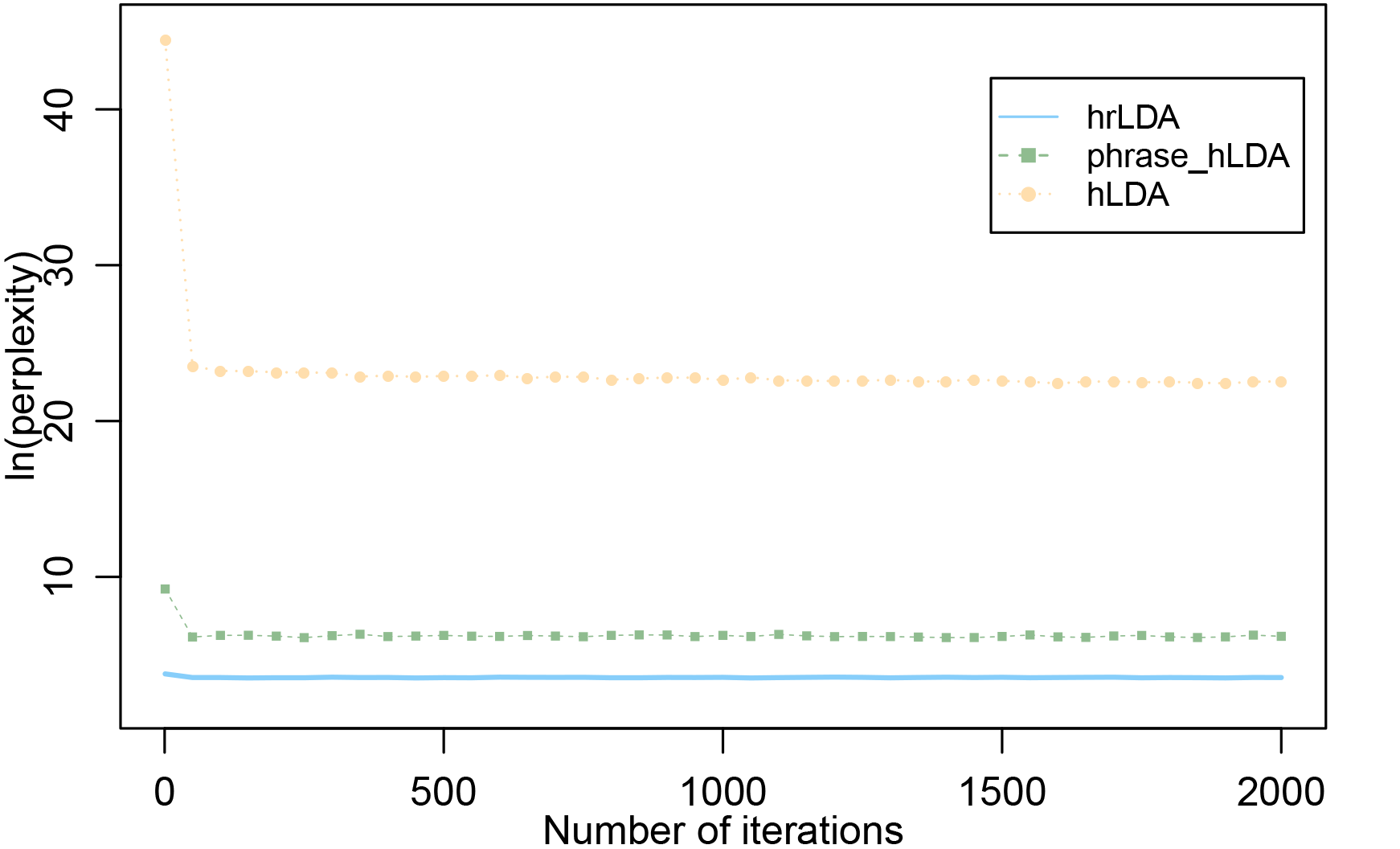}
  \caption{The entire corpus}
  \label{fig:entire corpus 6}%
\end{subfigure}
\caption{Perplexity trends within 2000 iterations with level = 6}
\label{fig: level 6}
\end{figure*}

% Figure 6 level = 10
\begin{figure*}[pt]
\begin{subfigure}[b]{0.5\textwidth}

  \includegraphics[width = \textwidth]{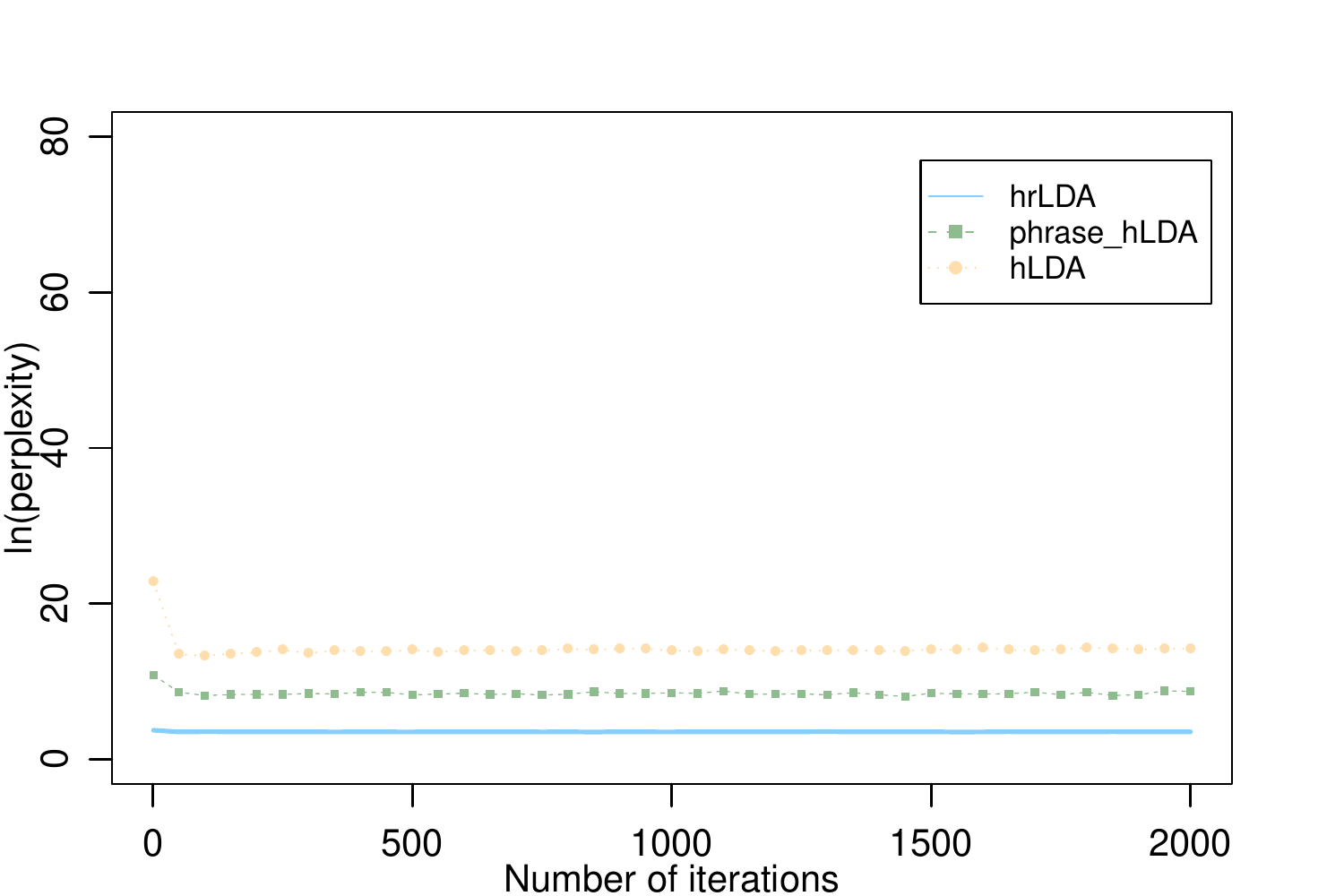}
  \caption{The Wiki corpus}
  \label{fig:Wiki corpus 10}
\end{subfigure}%
\begin{subfigure}[b]{0.5\textwidth}

  \includegraphics[width =\textwidth]{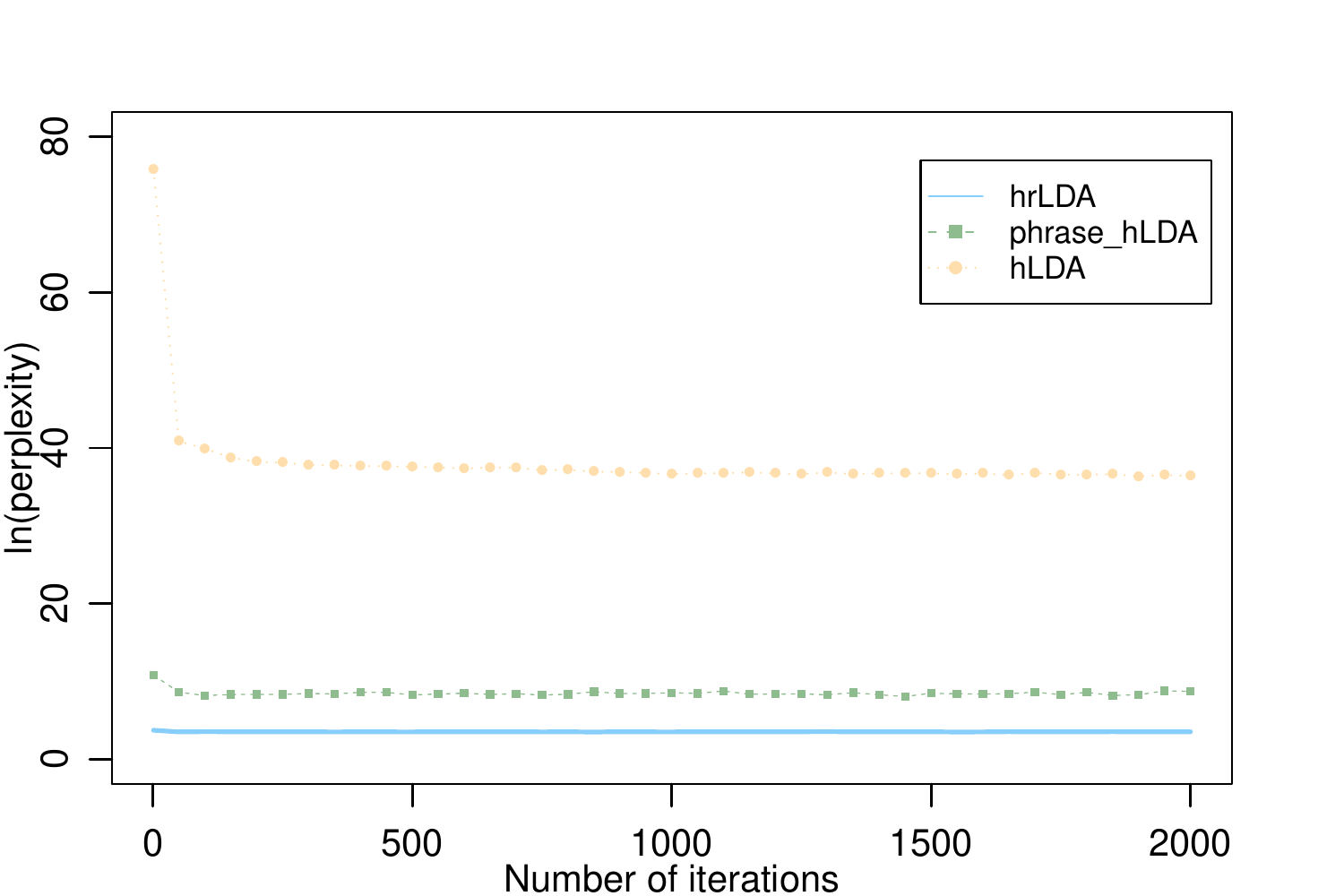}
  \caption{The entire corpus}
  \label{fig:entire corpus 10}%
\end{subfigure}
\caption{Perplexity trends within 2000 iterations with level = 10}
\label{fig: level 10}
\end{figure*}

% Figure 8 level = 10 final
\begin{figure*}[pt]
\begin{subfigure}[b]{0.5\textwidth}

  \includegraphics[width = \textwidth]{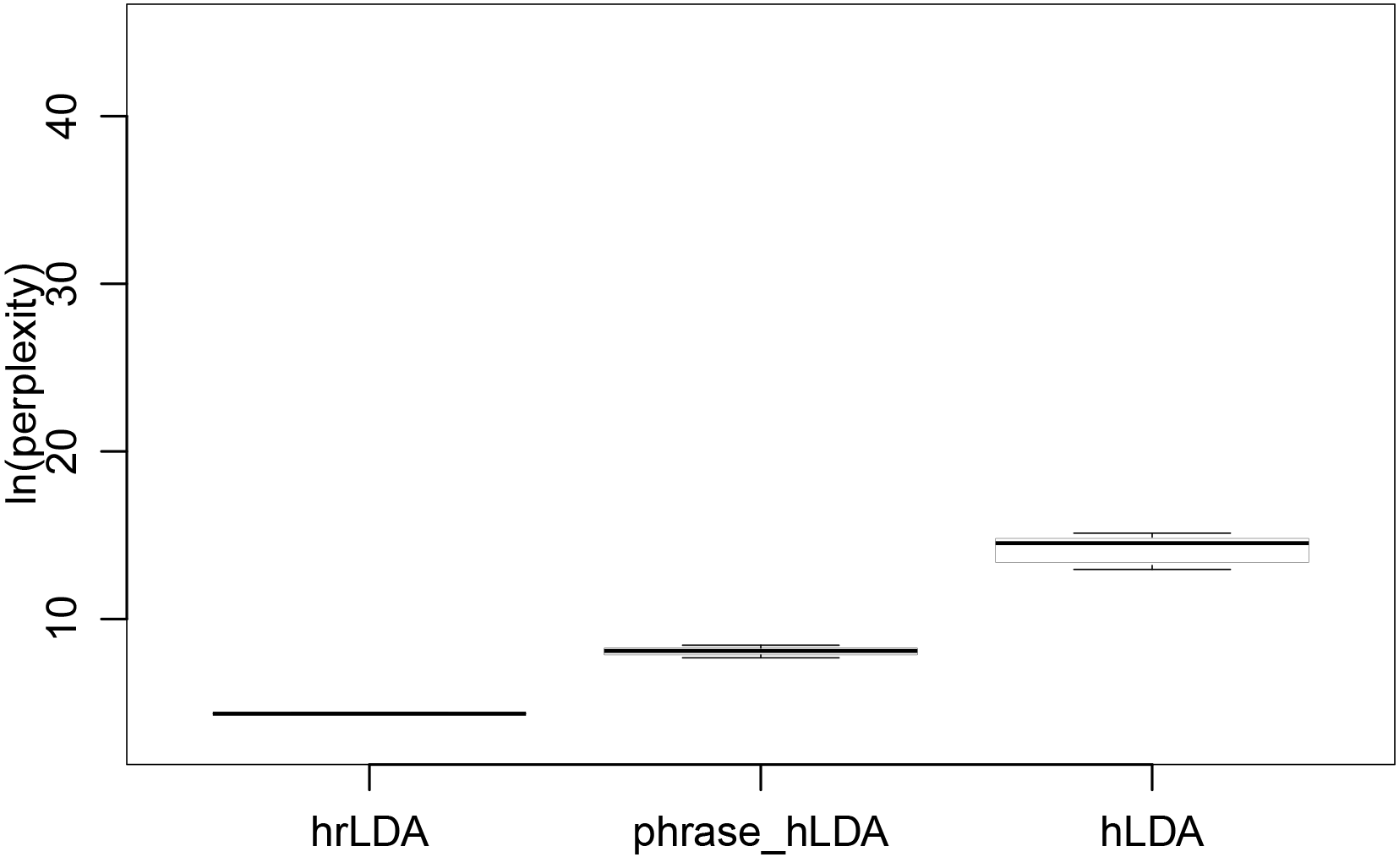}
  \caption{The Wiki corpus}
  \label{fig:Wiki corpus 10 final}
\end{subfigure}%
\begin{subfigure}[b]{0.5\textwidth}

  \includegraphics[width =\textwidth]{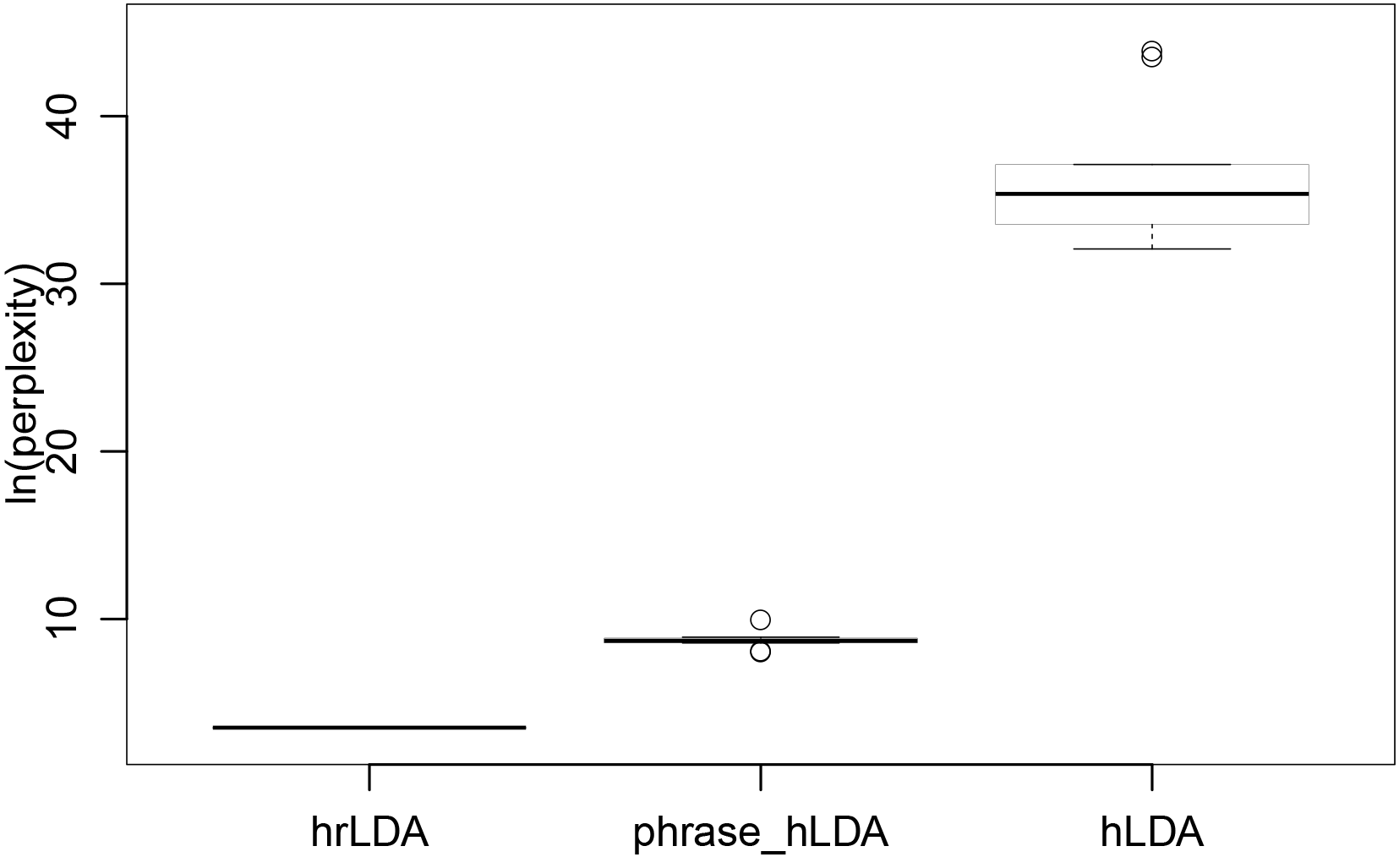}
  \caption{The entire corpus}
  \label{fig:entire corpus 10 final}%
\end{subfigure}
\caption{Average perplexities with confidence intervals of the three models in the final 2000th iteration with level = 10}
\label{fig: level 10 final}
\end{figure*}

Our interpretation is that hLDA and phrase\_hLDA tend to assign terms to the largest topic and thus do not guarantee that each topic path contains terms with similar meaning.  

\noindent\textbf{Robustness}

Figure \ref{fig: noisy results} shows exhaustive hierarchical topic trees extracted from a small text sample with topics from four domains: $semiconductor$, $integrated$ $circuit$, $Berlin$, and $London$. hLDA tends to mix words from different domains into one topic. For instance, words on the first level of the topic tree come from all four domains. This is because the topic path drawing method in existing hLDA-based models takes words in the most important topic of every document and labels them as the main topic of the corpus. In contrast, hrLDA is able to create four big branches for the four domains from the root. Hence, it generates clean topic hierarchies from the corpus. 

\begin{figure}[pt]
\begin{subfigure}[b]{0.5\textwidth}
  \includegraphics[width =\textwidth]{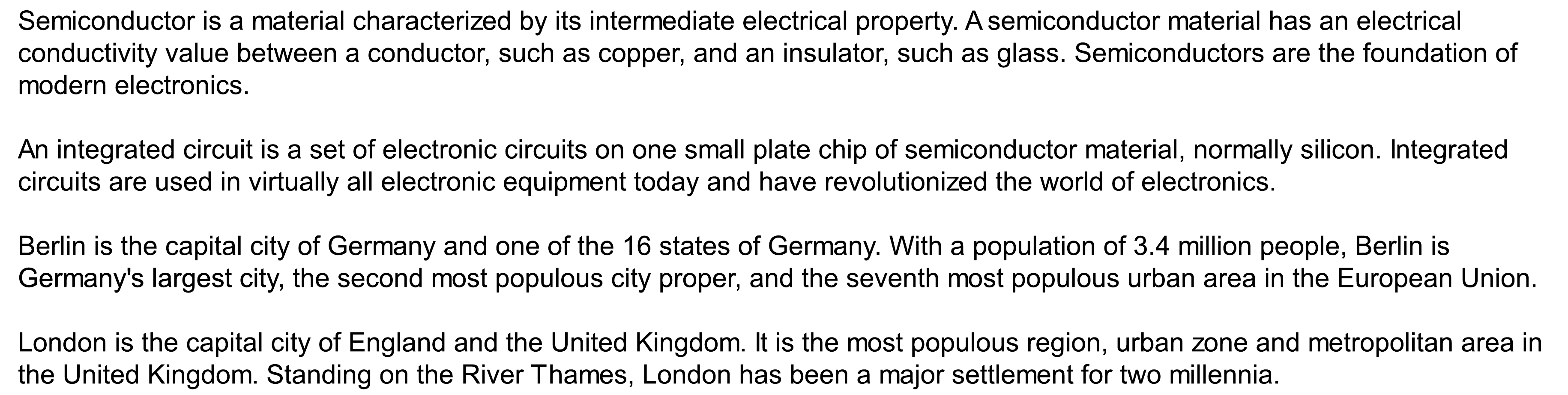}
  \caption{A toy corpus in domains: semiconductor, integrated circuit, Berlin, and London}
  \label{fig:noisy text}%
\end{subfigure}
\begin{subfigure}[b]{0.5\textwidth}
  \includegraphics[width =\textwidth]{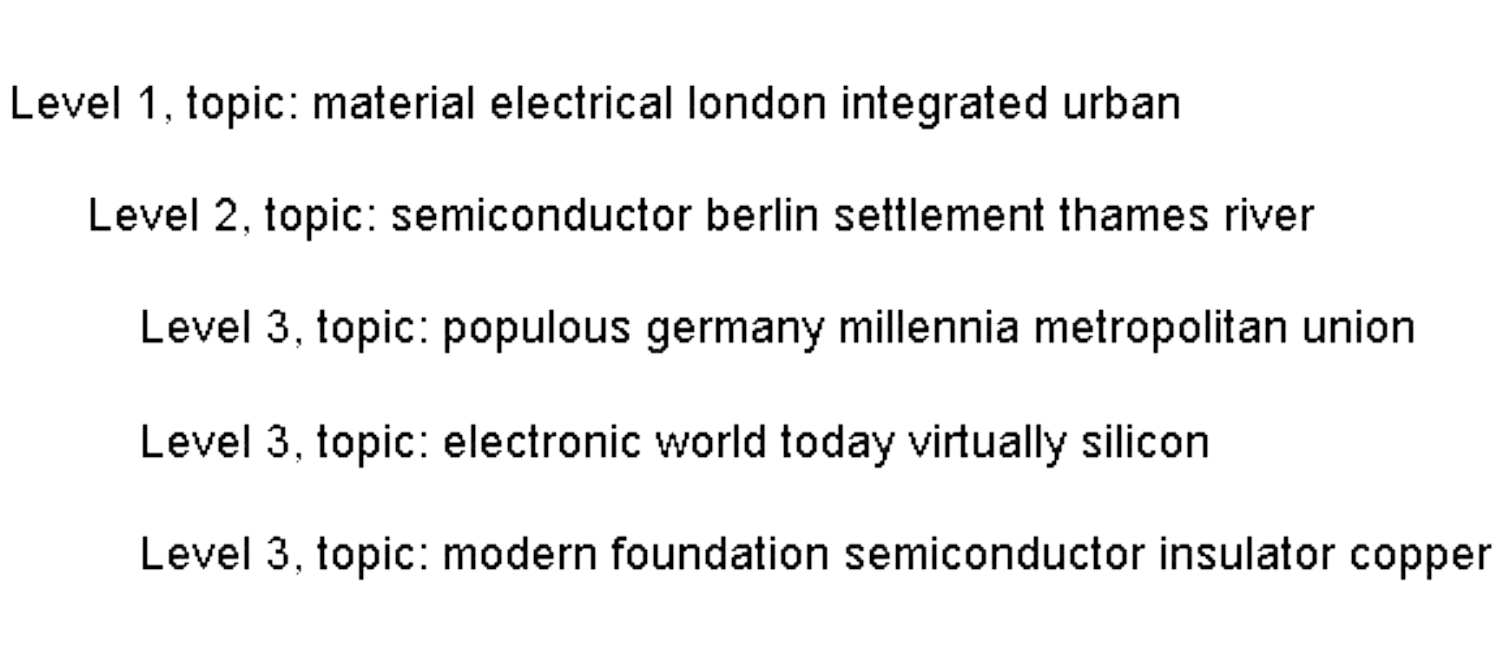}
  \caption{The topic tree obtained from hLDA; each node contains the top five words ordered by their probabilities of being in the corresponding topics}
  \label{fig:noisy hLDA result}%
\end{subfigure}
\begin{subfigure}[b]{0.5\textwidth}
  \includegraphics[width =\textwidth]{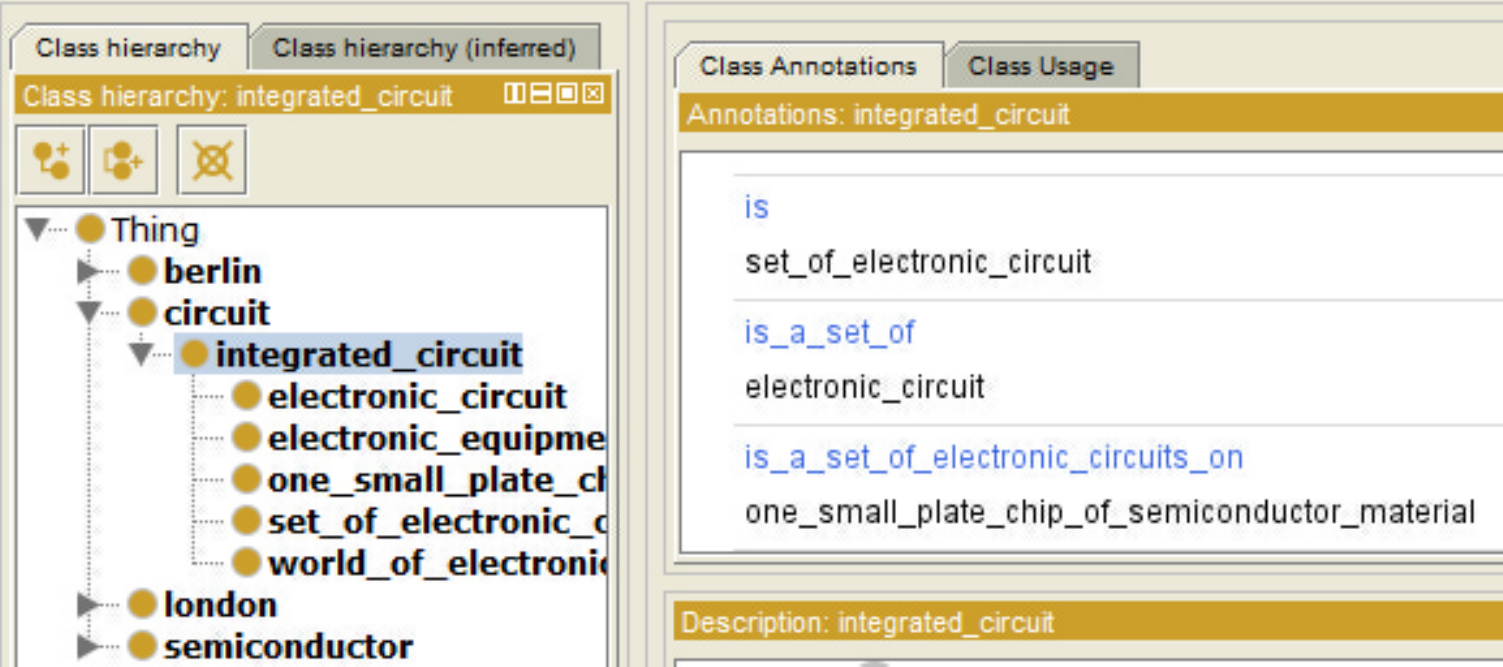}
  \caption{The topic tree (left panel \textit{class hierarchy}) with relations (right panel \textit{class annotations}) obtained from hrLDA}
  \label{fig:noisy hrLDA result}%
\end{subfigure}
\caption{Performance of hLDA and hrLDA on a toy corpus of diversified topics}
\label{fig: noisy results}
\end{figure}

\subsection{Gold Standard-based Ontology Evaluation}

The visualization of one concrete ontology on the $semiconductor$ $packaging$ domain is presented in Figure \ref{fig: ontology 10}. For instance, Topic \textit{packaging} contains topic \textit{integrated circuit packaging}, and topic label \textit{jedec} is associated with relation triplet \textit{(jedec, be short for, joint electron device engineering council)}.

%Figure 10
\begin{figure}[pt]
  \includegraphics[width =0.5\textwidth]{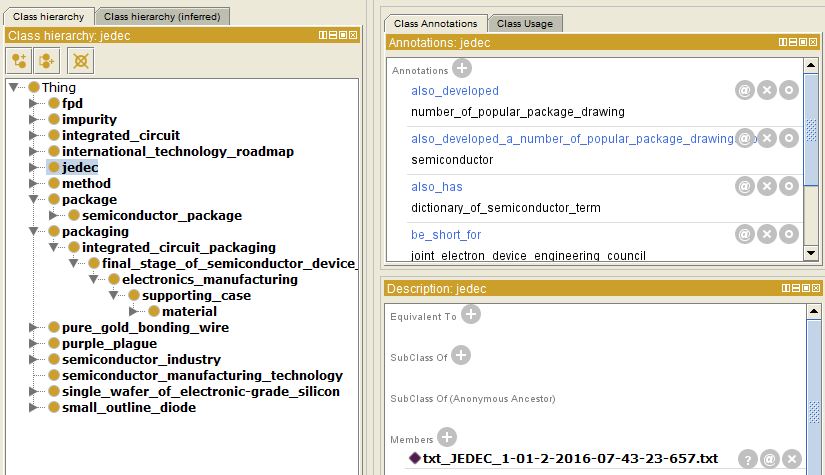}
  \label{fig:semiconductor ontology 10}
 \caption{A 10-level semiconductor ontology that contains 2063 topics and 6084 relation triplets}
\label{fig: ontology 10}
 
\end{figure}

We use KB-LDA, phrase\_hLDA, and LDA+GSHL as our baseline methods, and compare ontologies extracted from hrLDA, KB-LDA, phrase\_hLDA, and LDA+GSHL with DBpedia ontologies. We use precision, recall and F-measure for this ontology evaluation. A true positive case is an ontology rule that can be found in an extracted ontology and the associated ontology of DBpedia. A false positive case is an incorrectly identified ontology rule. A false negative case is a missed ontology rule. Table \ref{table:ar} shows the evaluation results of ontologies extracted from Wikipedia articles pertaining to \textit{European Capital Cities} (Corpus E), \textit{Office Buildings in Chicago} (Corpus O) and \textit{Birds of the United States} (Corpus B) using hrLDA, KB-LDA, phrase\_hLDA (tree depth $L$ = 3), and LDA+GSHL in contrast to these gold ontologies belonging to DBpedia. The three corpora used in this evaluation were collected from Wikipedia abstracts, the same text source of DBpedia. The seeds of hrLDA and the root concepts of LDA+GSHL are \textit{capital}, \textit{building}, and \textit{bird}. For both KB-LDA and phrase\_hLDA we kept the top five tokens in each topic as each node of their topic trees is a distribution/list of phrases. hrLDA achieves the highest precision and F-measure scores in the three experiments compared to the other models. KB-LDA performs better than phrase\_hLDA and LDA+GSHL, and phrase\_hLDA performs similarly to LDA+GSHL. In general, hrLDA works well especially when the pre-knowledge already exists inside the corpora. Consider the following two statements taken from the corpus on \textit{Birds of the United States} as an example. In order to use two short documents ``\textit{The Acadian flycatcher is a small insect-eating bird.}" and ``\textit{The Pacific loon is a medium-sized member of the loon.}" to infer that \textit{the Acadian flycatcher} and \textit{the Pacific loon} are both related to topic \textit{bird}, the pre-knowledge that ``\textit{the loon is a species of bird}" is required for hrLDA. This example explains why the accuracy of extracting ontologies from this kind of corpus is low.

% Table 1
\begin{table}[h]
\begin{center}
\caption{Precision, recall and F-measure (\%)}
\scalebox{1.0}{
\begin{tabular}{llrrr}
Domain &  &{Corpus E} &{Corpus O} &{Corpus B} \\
\hline							
							&hrLDA &\textbf{96.0} &\textbf{92.4} &\textbf{84.0} \\
							&KB-LDA &90.7  &89.9  &79.4  \\							
							&phrase\_hLDA &27.6 &27.4 &24.5 \\						
\multirow{-3}{*}{Precision}  &LDA+GSHL &52.4  &19.8  &28.6  \\				                            
\hline							
							&hrLDA &\textbf{86.9} &74.7 &\textbf{81.9} \\
							&KB-LDA &83.8  &\textbf{75.4}  &63.3  \\
							&phrase\_hLDA &50.6 &57.5 &36.5 \\				
\multirow{-3}{*}{Recall}     &LDA+GSHL &20.0  &73.1  &11.8  \\
\hline							
							&hrLDA &\textbf{91.2} &\textbf{82.6} &\textbf{82.9} \\
							&KB-LDA &87.1  &82.0  &70.4  \\
							&phrase\_hLDA &35.7 &26.8 &29.3 \\	                           
\multirow{-3}{*}{F-measure}  &LDA+GSHL &29.0  &31.2  &16.7  \\
\label{table:ar}		
\end{tabular}}
\end{center}
 
\end{table}

\section{Concluding Remarks}

In this paper, we have proposed a completely unsupervised model, hrLDA, for terminological ontology learning. hrLDA is a  domain-independent and self-learning model, which means it is very promising for learning ontologies in new domains and thus can save significant time and effort in ontology acquisition. 

We have compared hrLDA with popular topic models to interpret how our algorithm learns meaningful hierarchies. By taking syntax and document structures into consideration, hrLDA is able to extract more descriptive topics. In addition, hrLDA eliminates the restrictions on the fixed topic tree depth and the limited number of topic paths. Furthermore, ACRP allows hrLDA to create more reasonable topics and to converge faster in Gibbs sampling. 

We have also compared hrLDA to several unsupervised ontology learning models and shown that hrLDA can learn applicable terminological ontologies from real world data. Although hrLDA cannot be applied directly in formal reasoning, it is efficient for building knowledge bases for information retrieval and simple question answering. Also, hrLDA is sensitive to the quality of extracted relation triplets. In order to give optimal answers, hrLDA should be embedded in more complex probabilistic modules to identify true facts from extracted ontology rules. Finally, one issue we have not addressed in our current study is capturing pre-knowledge. Although a direct solution would be adding the missing information to the data set, a more advanced approach would be to train topic embeddings to extract hidden semantics.

\section*{Acknowledgments}
This work was supported in part by Intel Corporation, Semiconductor Research Corporation (SRC). We are obliged to Professor Goce Trajcevski from Northwestern University for his insightful suggestions and discussions. This work was partly conducted using the Protege resource, which is supported by grant GM10331601 from the National Institute of General Medical Sciences of the United States National Institutes of Health.

%------------------------------------------------------------------------- 
\bibliographystyle{IEEEtran}
\bibliography{IEEEabrv,hrLDA_IEEE_IRI_17}

\end{document}